# Development and validation of a natural language processing algorithm to pseudonymize documents in the context of a clinical data warehouse

Authors and affiliations:

**Xavier Tannier, PhD (ORCID: 0000-0002-2452-8868):** Sorbonne Université, Inserm, Université Sorbonne Paris Nord, Laboratoire d'Informatique Médicale et d'Ingénierie des Connaissances pour la e-Santé (LIMICS), Paris, France
**Perceval Wajsbürt, PhD (ORCID: tbc.):** Innovation and Data unit, IT Department, Assistance Publique-Hôpitaux de Paris, Paris, France
**Alice Calliger (ORCID: 0000-0002-5767-313X):** Innovation and Data unit, IT Department, Assistance Publique-Hôpitaux de Paris, Paris, France
**Basile Dura (ORCID: 0000-0002-8315-4050):** Innovation and Data unit, IT Department, Assistance Publique-Hôpitaux de Paris, Paris, France
**Alexandre Mouchet:** Innovation and Data unit, IT Department, Assistance Publique-Hôpitaux de Paris, Paris, France
**Martin Hilka, PhD:** Innovation and Data unit, IT Department, Assistance Publique-Hôpitaux de Paris, Paris, France
**Romain Bey, PhD (ORCID: 0000-0002-6413-5188):** Innovation and Data unit, IT Department, Assistance Publique-Hôpitaux de Paris, Paris, France

Corresponding author:
Xavier Tannier
Mailing address:
LIMICS
15, rue de l'école de médecine
75006 PARIS
FRANCE

Phone number: +33 1 44 27 91 13

# Abstract

The objective of this study is to address the critical issue of de-identification of clinical reports in order to allow access to data for research purposes, while ensuring patient privacy. The study highlights the difficulties faced in sharing tools and resources in this domain and presents the experience of the Greater Paris University Hospitals (AP-HP) in implementing a systematic pseudonymization of text documents from its Clinical Data Warehouse.

We annotated a corpus of clinical documents according to 12 types of identifying entities, and built a hybrid system, merging the results of a deep learning model as well as manual rules. Our results show an overall performance of 0.99 of F1-score.

We discuss implementation choices and present experiments to better understand the effort involved in such a task, including dataset size, document types, language models, or rule addition. We share guidelines and code under a 3-Clause BSD license.

# I. Introduction

De-identification of textual clinical reports consists of removing or replacing protected health information (PHIs) from the electronic health reports (EHRs), in order to limit the risk of recognizing a patient for someone who would not be part of the care team. Although the definition of PHIs and the regulatory constraints may vary according to the country and the situation, de-identification is an essential step in order to allow access to data for research purposes, in France as in many other countries. Research projects involving identifying documents is only possible with an informed consent from the patients concerned, which is impracticable in studies involving thousands or even millions of patients, or with an hypothetical waiver of informed consent from the institutional review board (IRB).

The de-identification of clinical reports is a critical issue since the automatic analysis of clinical reports using natural language processing (NLP) algorithms is a cornerstone of EHR studies, often in a multicenter perspective.

A lot of work has been done on this topic, in several languages [1,2,3], including French [4,5,6,7,8]. Different scenarios have been proposed to improve the processing of this task [4, 9]. Yet, there is no consensus method or protocol in the community, and more importantly it is very difficult for new actors to benefit from the experience and tools implemented by others, for several reasons:

- **Language**: De-identification is a very language-dependent process. It is not easy to adapt tools made for other languages.
- **Data sharing**: While rule-based approaches, which are easily shareable, have been proposed, their quality has been quickly surpassed by statistical learning tools, requiring the annotation of a training corpus. However, by definition, a corpus

- annotated for this task cannot be shared. Even with a replacement of the identifying features by fake surrogates (pseudonymization), the security obtained is not considered sufficient, at least according to the European regulation, for an open sharing.
- **Model sharing**: Models trained by supervised learning methods are themselves not shareable, as the impossibility of recovering portions of the original data by attacking the model is not proven [10, 11].
- **Interoperability**: Personal metadata (last name, first name, address, date of birth) are generally available in the structured information of the clinical records. Although insufficient to de-identify a text by simple mapping, the use of this metadata in a hybrid way generally improves the performance significantly [5, 12]. However, the heterogeneous formats of this data between services or hospitals do not help sharing.
- **Performance**: Finally, the performance requirements are very high for this task, especially in terms of recall (sensitivity) in order to sufficiently reduce the risk of re-identification. Transfer learning approaches for language or domain adaptation [13], distant supervision, low-data settings, while very promising in the general setting, are accompanied by a significant drop in performance, not acceptable in this context.

The Greater Paris University Hospitals (AP-HP for *Assistance Publique - Hôpitaux de Paris*) have implemented as of 2019 a systematic pseudonymization of text documents [5] within its Clinical Data Warehouse (CDW) in OMOP format (Observational Medical Outcomes Partnership [14]). This process has been updated and modernized very recently.

In this article, we relate the experience of AP-HP in this matter. We describe the implemented system, but also all the resources (code, lists) that we were able to share at the end of this work. We also discuss our main implementation choices, the size of the training corpus, the types of documents included in the dataset, the interest of fine-tuning a specific language model or adding static or dynamic rules, as well as considerations on pre-processing, computational cost, carbon footprint and the specificities of medical documents for de-identification. We thus share a certain number of conclusions allowing to best size the human and machine effort to achieve an efficient de-identification system.

## II. Methods

We consider here de-identification by pseudonymization, as recommended by the independent French administrative regulatory body for the use of personal data (*Commission nationale de l'informatique et des libertés*, CNIL). Pseudonymization requires that the identifying data is replaced by a plausible surrogate that cannot be associated with the data without knowing a certain key, making it difficult to reestablish a link between the individual and their data. While not sufficient for fully open sharing, pseudonymization makes false negatives more difficult to detect when entities are substituted, and allows data to be accessed in a secure space for research purposes. An automatic, strict anonymization, i.e., the irreversible and flawless removal of the link between the individual and his or her medical record, must still be considered impossible for textual records at this time.

The types of identifying characteristics that we considered are listed at Table 1.

| Label | Description |
| --- | --- |
| ADDRESS | Street address, eg., 33 boulevard de Picpus |
| DATE | Any absolute date other than a birthdate |
| BIRTHDATE | Birthdate |
| HOSPITAL | AP-HP Hospital: not replaced but prevents false positive names or cities |
| PATIENT ID | Any internal AP-HP identifier for patients, displayed as a number |
| EMAIL | Any email address |
| VISIT ID | Any internal AP-HP identifier for visits, displayed as a number |
| LASTNAME | Any last name (patients, doctors, third parties, etc.) |
| FIRSTNAME | Any first name (patients, doctors, third parties, etc.) |
| SSN | Social security number |
| PHONE | Any phone number |
| CITY | Any city |
| ZIP CODE | Any zip code |

Table 1: List of identifying characteristics considered in this work

## Data

We selected a set of EHRs split between training, development and test sets. The documents were annotated in two phases by 3 annotators and reviewed by 3 other annotators to ensure the consistency of annotations. Training documents were randomly sampled from post-2017 AP-HP medical records to bias the distribution and better fit the model to more recent documents. Test documents were sampled without constraint to evaluate performance on all documents in the CDW. Annotated entities are listed and described in Table 1. Documents were first pre-annotated with the predictions of a former pseudonymisation system [5]. Annotator agreement is measured both in terms of exact entity matching and token matching using micro-average F1-score. F1-score is the unweighted harmonic mean between precision (positive predictive value) and recall (sensitivity).

The texts were extracted from PDF files using two methods illustrated in Figure 1:
- **pdfbox**: a classic method adapted from the Apache PDFBox tool[1] with a defective rule-based layout segmentation that, in effect, extracts nearly all the text from the PDF
- **edspdf**: a new segmentation-based PDF body text extraction, implemented by the APHP-CDW team[2]

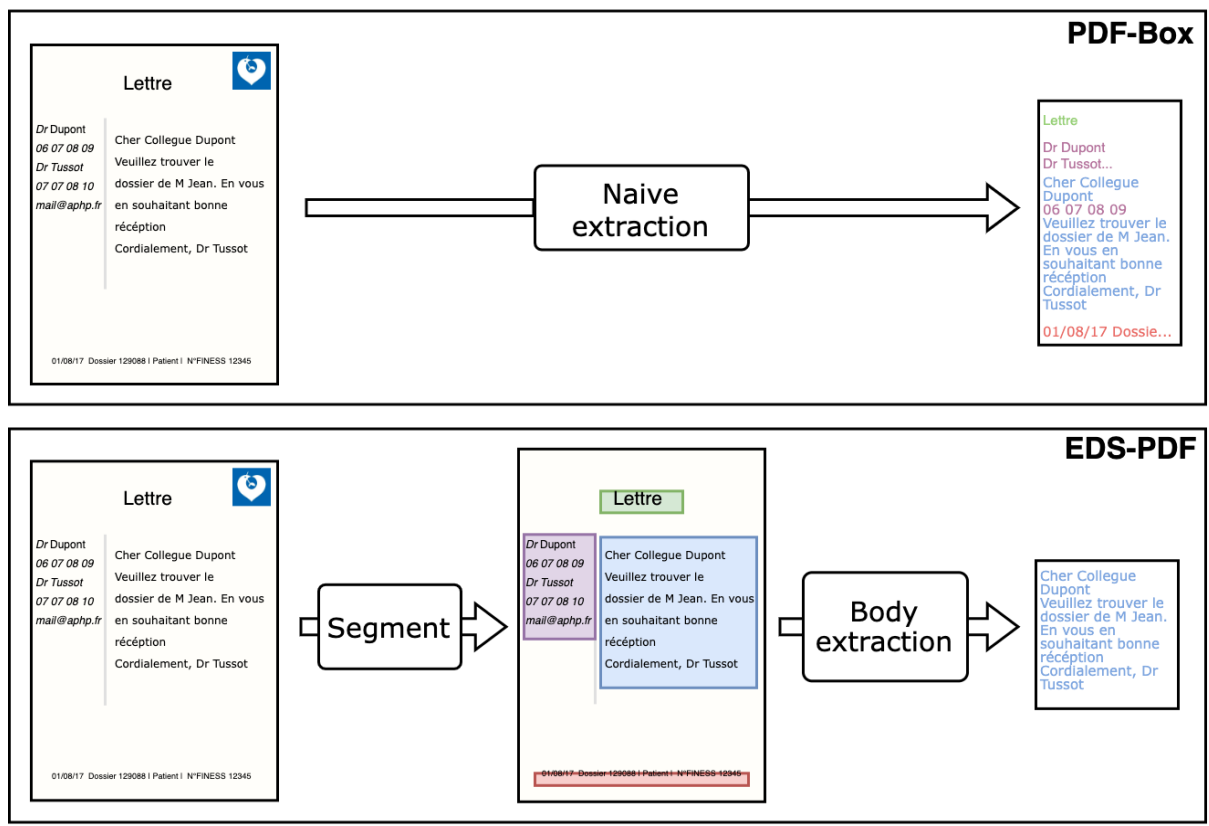

Figure 1: Overview of the two PDF extraction methods used to generate texts in the dataset

In order to evaluate the accuracy of our pseudonymisation pipeline in both setups, we annotated the extraction of each method for each document of the test set. We annotated 90% of the training data using the edspdf method and 10% using the pdfbox method for robustness purposes. This choice was made in order to ensure the robustness of the system to environments not using an elaborate extraction tool, as well as to guarantee the presence of a significant number of identifying entities in the training set.

---

[1] https://pdfbox.apache.org/
[2] Article in the process of being submitted.

Subjects that objected to the reuse of their data were excluded. This study was approved by the local institutional review board (IRB00011591, decision CSE22-19).
As the data for the outcomes are routinely collected, we follow the REporting of Studies Conducted using the Observational Routinely Collected Health Data (RECORD) statement [15] (see Supplementary Material A).

## Preprocessing

Documents are tokenized into words using the spaCy-based EDS-NLP framework (version 0.7.3) [16]. This tokenization affects both the training of the machine learning model, the granularity of the rule-based approach and the scoring of these models. For instance, intra word entities (e.g., "JamesSmith") were merged and labeled as the last entity.
Each document was also enriched with some metadata of the patient when available, such as names, birthdates, city of residence and so forth. During training, long documents were segmented into several samples of up to 384 words, cutting as needed at the beginning of each sentence, as detected by a heuristic.

## Rule-based approach

The rule-based approach focuses on achieving higher precision than recall. Indeed, since the predicted entities of the rule-based and trained systems are merged (and not crossed for instance) to maximize the recall of the hybrid model, the false positives of the rule-based system cannot be recovered.

Our rule-based approach uses both static rules based on regular expressions (regexes) and dynamic metadata-based rules when applicable. These rules are coarsely described in Table 2.

| Label | Static rule | Dynamic rule |
| --- | --- | --- |
| DATE | regex | |
| BIRTHDATE | DATE with birth pattern nearby | DATE matching birthdate |
| HOSPITAL | gazetteer | |
| PATIENT ID | | lowercase matching |
| EMAIL | regex | lowercase matching |
| VISIT ID | regex | lowercase matching |
| PHONE | regex | strict matching |
| SSN | regex | strict matching |

| LASTNAME | regex (looking for a combination of these and common prefixes such as Mr./Mrs./Dr.) | strict matching |
|---|---|---|
| FIRSTNAME | | strict matching |
| ADRESSE | regex (looking for a combination of these entities to avoid false positives) | lowercase matching |
| CITY | | strict matching |
| ZIP CODE | | strict matching |

Table 2: Static and dynamic rules used to supplement the predictions of the machine learning model

## Machine-learning approach

The machine learning approach uses a standard token classification model for general-purpose named entity recognition (NER) composed of a Transformer model [17] followed by a stack of constrained Conditional Random fields (CRF) classification heads [18]. The model is briefly described in Figure 2, and detailed in [19].

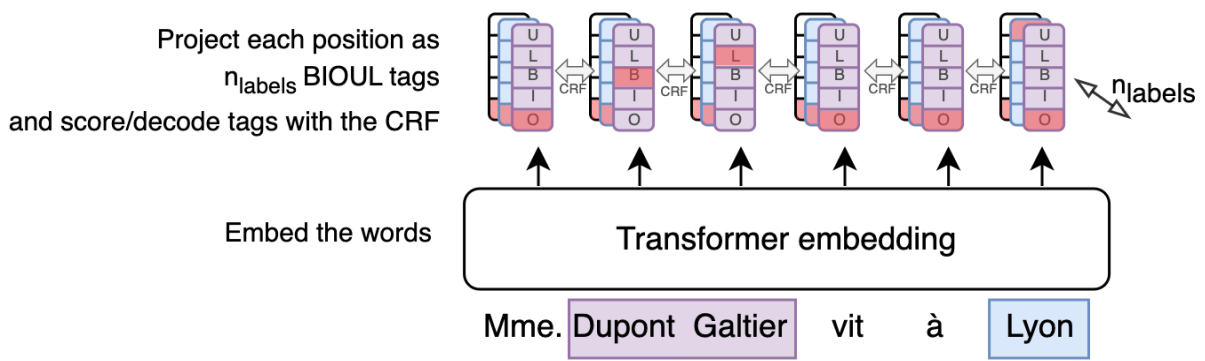

Figure 2: Overview of the named entity recognition model used in the de-identification system.

We performed several experiments with different Transformer models, either pre-trained or fine-tuned from general domain documents or CDW documents (as exported by the legacy PDF processing pipeline "pdfbox") following [20]:
- the "*camembert base*" model, pre-trained on texts from the general domain
- "*fine-tuned*": a fine-tuned model, initialized with "camembert base" weights and trained for 250 000 steps with the Whole Word Masked Language modeling task, on non-pseudonymised data
- "*scratch pseudo*": The model pre-trained from scratch on pseudonymised CDW data, from [20]

## Hybrid approach

In the hybrid approach, we run both the machine learning system and the rule-based system and merge the output of both. To maximize the masking of identifying information (i.e., recall), we output an entity if it is predicted by at least one of the two systems. In the case of overlapping entities, we chose the largest one, and the deep learning prediction when they overlap perfectly but the type differs.

To ensure optimal performance, we kept all the rules that did not lower the precision of the model as determined through evaluation on the development set, i.e. all the rules from Table 2 except static FIRSTNAME, LASTNAME and DATE.

## Entity replacement

Entities that are detected are not deleted or masked. Instead, they are replaced by plausible substitutes that maintain the plausibility of the texts across a whole patient file (e.g., using the same name for the same patient). Entities are normalized first (by removing spaces and making them lowercase) before choosing a replacement. Dates are also replaced, with a random but consistent shift for the same patient, which allows to keep the right temporal distances between events. Replacements and date shifts are recalculated for each new cohort extraction for research projects in order to avoid any risk of crossing information between cohorts (see Figure 3).

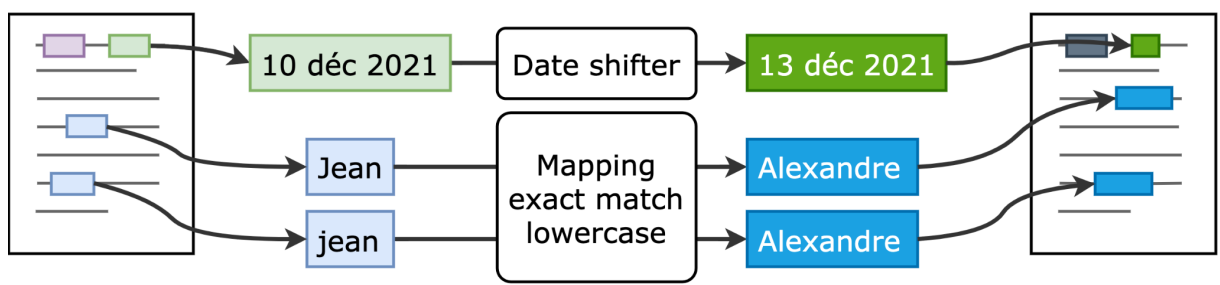

Figure 3: Overview of the entity replacement.

## Experimental setups

In order to obtain generalizable lessons on this work, we implemented the following additional experiments:

- **Impact of the size of the training dataset**: we trained the NER model using different sizes of training set to determine the optimal annotation effort to provide. We showed the performance of the obtained model on each entity type.
- **Impact of document types**: We estimated the influence of the different types of documents on the generalizability of the system, by ablating certain types of documents from the training set and evaluating the performance of the model on these types in the test set. We chose 5 types that are not the most frequent but

- nevertheless significantly present in the warehouse, for this ablation study: Pathology notes (CR-ANAPATH), Diagnostic study notes (CR-ACTE-DIAG-AUTRE), Multidisciplinary team meetings (RCP), Diagnostic imaging studies (CR-IMAGE), Surgery notes (CR-OPER) (see the distribution of these types in Figure 4).
- **Impact of the language model**: We evaluated whether fine-tuning or retraining a language model on the CDW was relevant, by using a general-domain CamemBERT model [19], a CamemBERT model fine-tuned, or retrained, with the CDW data [20].
- **Impact of the text extraction step:** We compared how the performance of the model was affected by the PDF extraction preprocessing step. In particular, we measured the number of fully redacted documents given that the "edspdf" method already removes margins and headers that contain much identifying information.
- We also presented results in terms of inference time and carbon footprint.

We evaluated our model through several commonly used metrics including precision (positive predictive value), recall (sensitivity), and F1-score (harmonic mean between precision and recall). Additionally, we introduced two novel metrics, "redacted" and "fully redacted", to provide a more comprehensive evaluation of the model's performance. "Redacted" measures the recall of the model at the token-level, regardless of the token label. This takes into account the fact that a type error can still remove identifying information. On the other hand, "Fully redacted" evaluates the percentage of documents where all entities have been correctly redacted, i.e., with a perfect token-level recall score regardless of the labels and a perfectly de-identified report.

## III. Results

### Data

We produced a manually annotated dataset made of 3682 documents for training and evaluating the process. The numbers of annotated entities per entity type, per text-extraction method and per split (i.e., training/development/test sets) are described in Table 3.
Considering only documents whose texts were extracted by the edspdf method, Figure 4 details for each document type the average per-document numbers and types of identifying entities and supplementary results B shows the distribution of the total number of annotated entities by document type.

The average inter-annotator agreement is above 0.8 for all entities, except for the hospital names, which proved to be ambiguous during the first annotation step (ambiguities with the clinic names for example). The Supplementary Results C provide details on the inter-annotator agreements by entity types.

|  | Training set | | Development set | | Test set | |
| --- | --- | --- | --- | --- | --- | --- |
|  | edspdf | pdfbox | edspdf | pdfbox | edspdf | pdfbox |
| **ADDRESS** | 212 | 543 | 10 | 35 | 12 | 625 |
| **BIRTHDATE** | 916 | 519 | 52 | 31 | 87 | 484 |
| **CITY** | 592 | 742 | 27 | 47 | 44 | 810 |
| **DATE** | 14711 | 2360 | 878 | 113 | 1973 | 2831 |
| **EMAIL** | 20 | 182 | 0 | 17 | 1 | 166 |
| **FIRSTNAME** | 3468 | 3826 | 215 | 216 | 478 | 3739 |
| **HOSPITAL** | 1451 | 758 | 87 | 47 | 162 | 796 |
| **LASTNAME** | 4910 | 4299 | 292 | 236 | 625 | 4150 |
| **NSS** | 73 | 79 | 6 | 7 | 4 | 32 |
| **PATIENT ID** | 121 | 339 | 8 | 18 | 8 | 392 |
| **PHONE** | 397 | 1589 | 23 | 148 | 77 | 1851 |
| **VISIT ID** | 49 | 283 | 7 | 17 | 6 | 282 |
| **ZIP** | 215 | 552 | 10 | 35 | 14 | 635 |
| **ENTS** | 27135 | 16071 | 1615 | 967 | 3491 | 16793 |
| **Documents** | 3025 | 348 | 200 | 22 | 348 | 348 |

Table 3: Number of identifying entities composing the training set, the development set and the test set, divided per entity type and per method used for the extraction of text from PDF files.

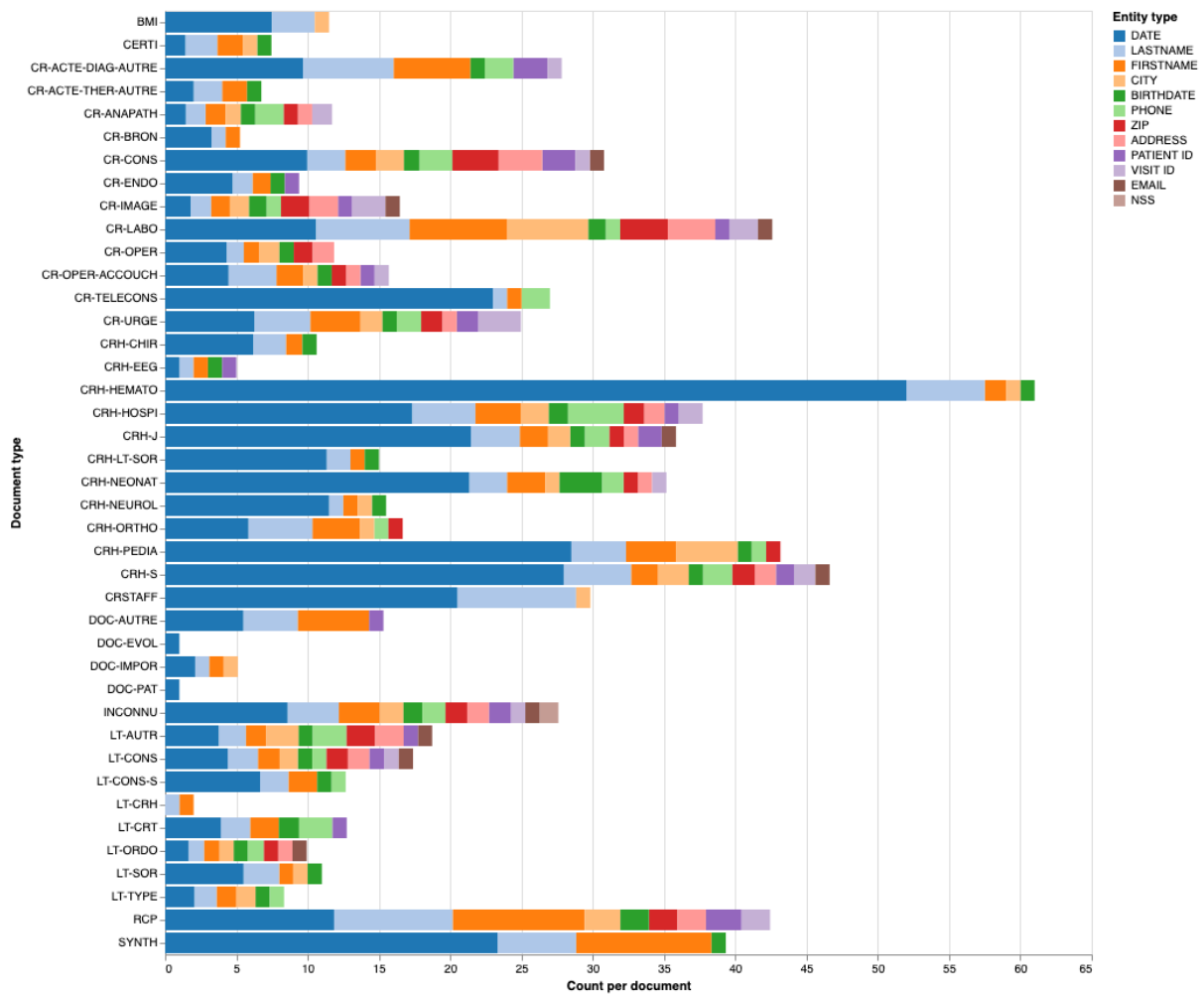

Figure 4: Average number of identifying entities found in an annotated document for each document type and subdivided per entity type. Only documents whose texts were extracted by the edspdf method were considered. The prefix "CR" in the Document Type stands for Report and the prefix "LT" for Letter.

## De-identification performance

Table 4 shows performances of de-identification for each entity type, and compares the results obtained by rules only, machine-learning only and both combined. We present in Supplementary Results D the performance obtained by the rule-based system on the development set, in order to show how we selected rules that were used in the hybrid model.

|  | Precision | | | Recall | | | F1 | | |
|---|---|---|---|---|---|---|---|---|---|
| Model | RB | ML | Hybrid | RB | ML | Hybrid | RB | ML | Hybrid |
| ADDRESS | 99.2 | 99.6 | 99.0 | 78.5 | 98.1 | 98.4 | 87.7 | 98.9 | 98.7 |
| BIRTHDATE | 97.9 | 98.0 | 98.2 | 73.3 | 98.2 | 98.2 | 83.8 | 98.1 | 98.2 |
| CITY | 94.8 | 98.3 | 98.0 | 47.4 | 98.7 | 98.8 | 63.2 | 98.5 | 98.4 |
| DATE | 93.6 | 99.7 | 99.7 | 95.8 | 99.3 | 99.3 | 94.7 | 99.5 | 99.5 |
| EMAIL | 99.8 | 99.1 | 98.9 | 95.2 | 99.3 | 99.9 | 97.5 | 99.2 | 99.4 |
| FIRSTNAME | 96.7 | 98.8 | 98.8 | 36.9 | 98.4 | 98.4 | 53.4 | 98.6 | 98.6 |
| LASTNAME | 87.8 | 98.6 | 98.6 | 59.0 | 98.3 | 98.6 | 70.6 | 98.4 | 98.6 |
| NSS | 97.3 | 88.3 | 88.0 | 95.1 | 96.9 | 98.9 | 96.2 | 92.3 | 93.1 |
| PATIENT ID | 99.4 | 99.0 | 99.0 | 84.8 | 94.0 | 94.0 | 91.5 | 96.4 | 96.4 |
| PHONE | 99.9 | 99.6 | 99.6 | 95.2 | 99.6 | 99.7 | 97.5 | 99.6 | 99.7 |
| VISIT ID | 98.2 | 91.5 | 91.5 | 87.1 | 89.1 | 89.4 | 92.3 | 90.2 | 90.4 |
| ZIP | 100.0 | 99.9 | 99.9 | 81.2 | 99.3 | 99.9 | 89.6 | 99.6 | 99.9 |
| ALL | 95.6 | 99.1 | 99.0 | 80.5 | 98.8 | 98.9 | 87.4 | 99.0 | 99.0 |

|  | Redacted | | | Fully redacted | | |
|---|---|---|---|---|---|---|
| Model | RB | ML | Hybrid | RB | ML | Hybrid |
| ADDRESS | 79.8 | 98.3 | 98.5 | 83.9 | 98.0 | 98.4 |
| BIRTHDATE | 98.5 | 99.8 | 99.8 | 98.7 | 99.7 | 99.7 |
| CITY | 47.4 | 98.8 | 98.8 | 61.4 | 98.1 | 98.2 |
| DATE | 96.1 | 99.6 | 99.6 | 76.7 | 95.4 | 95.4 |
| EMAIL | 95.2 | 99.3 | 99.9 | 98.7 | 99.6 | 99.9 |
| FIRSTNAME | 43.7 | 99.4 | 99.4 | 46.7 | 97.4 | 97.4 |
| LASTNAME | 59.5 | 99.3 | 99.6 | 47.3 | 96.4 | 97.2 |
| NSS | 95.1 | 99.1 | 100.0 | 99.7 | 99.9 | 100.0 |
| PATIENT ID | 84.8 | 98.2 | 98.2 | 93.1 | 99.1 | 99.1 |
| PHONE | 95.2 | 99.6 | 99.7 | 93.2 | 98.8 | 99.0 |
| VISIT ID | 87.4 | 90.0 | 90.4 | 97.0 | 98.1 | 98.3 |
| ZIP | 81.3 | 99.3 | 99.9 | 87.4 | 99.3 | 99.9 |
| ALL | 83.0 | 99.4 | 99.4 | 31.9 | 84.4 | 86.2 |

Table 4: De-identification performances on the test set.

## Impact of the size of the training dataset

Figure 5 shows the performance of the model on the test set for varying numbers of documents sampled from the training set, from 10 up to 3373 examples (the entire training set). We observe that the performance of the model in terms of micro-average recall saturates around 1500 documents in the training set. On the other hand, the performance in terms of full identifying word recall is much more sensitive to the presence of error and we observe an improvement until 3000 documents for the fully redacted metric (see Supplementary Results E). This shows the importance of evaluating the performance of a model according to several metrics. More details about the impact of the size of the training set can be found in Supplementary Results E.

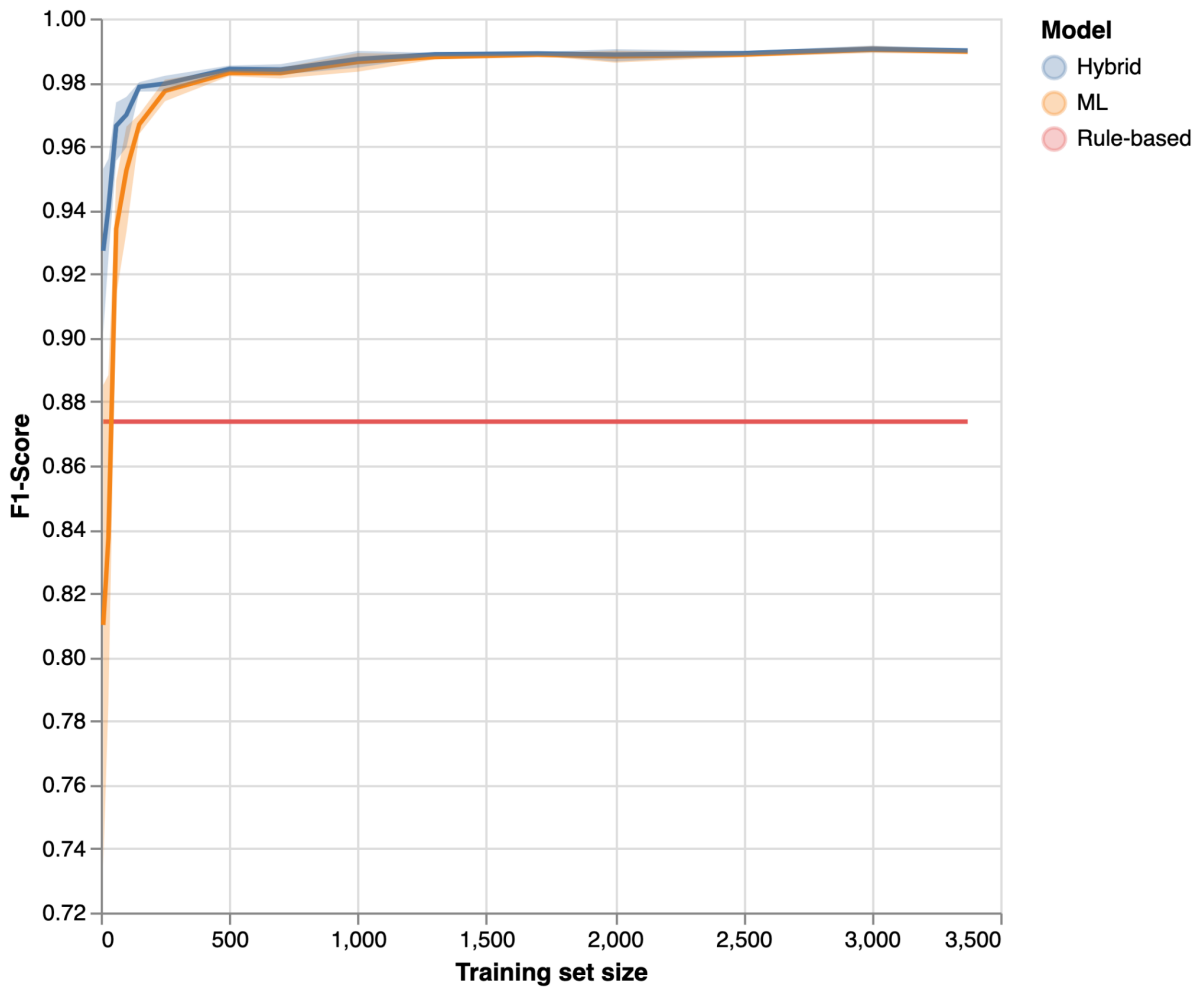

Figure 5: Performance of the model on the test set in token level F1-score for varying numbers of documents in the training set.

## Impact of the document types

In this experiment, the impact of removing certain types of documents from the training set was evaluated. Table 5 shows the performance of the de-identification on five document types, when each document type is excluded from or included in the training set. A Student t-test between the "Included" and "Excluded" model for each document type did not show any statistical significant difference in F1-score performance between these pairs of models, except in the case of "ANAPATH" documents (F1-score p-value = 0.027 < 0.05 and Fully redacted p-value = 0.023 < 0.05).

|  | F1-score | | Fully redacted | |
|---:|---:|---:|---:|---:|
| Doc type | Excluded | Included | Excluded | Included |
| Anapath | 95.9 ± 0.3 | 96.7 ± 0.5 | 66.7 ± 6.1 | 77.8 ± 5.0 |
| Misc diagnosis | 98.5 ± 0.8 | 98.3 ± 0.6 | 65.0 ± 14.3 | 65.0 ± 6.2 |
| MDM | 99.0 ± 0.3 | 99.0 ± 0.5 | 66.0 ± 4.9 | 78.0 ± 9.8 |
| Image | 96.6 ± 1.5 | 98.0 ± 1.0 | 94.5 ± 2.6 | 96.7 ± 1.5 |
| Post operative | 99.4 ± 0.1 | 99.4 ± 0.1 | 75.0 ± 0.0 | 75.0 ± 0.0 |

Table 5: Performances of the de-identification (F1-score and "fully redacted" metric) on five entity types, when each entity type is excluded from or included in the training set.

## Impact of the language model

| Transformer | Precision | Recall | F1-score | Redacted | Fully redacted |
|---|---:|---:|---:|---:|---:|
| fine-tuned | 97.8 ± 0.2 | 97.7 ± 0.2 | 97.8 ± 0.2 | 98.2 ± 0.2 | 75.5 ± 1.8 |
| camembert base | 96.8 ± 0.5 | 96.9 ± 0.1 | 96.8 ± 0.3 | 97.4 ± 0.1 | 68.9 ± 0.5 |
| scratch pseudo | 97.3 ± 0.1 | 97.2 ± 0.1 | 97.3 ± 0.1 | 97.6 ± 0.1 | 69.0 ± 1.0 |

Table 6: Performances of the de-identification according to the underlying language model

Table 6 shows the overall, "machine-learning only" performances obtained with either the general-domain CamemBERT model [19], our fine-tuned language model, or our CamemBERT model totally re-trained from scratch using the CDW data [20]. The "fine-tuned" transformer performs statistically better than its two counterparts by each metric.

## Impact of the PDF extraction step

Table 7 shows the impact of the method used to extract texts from PDF files (i.e., pdfbox or edspdf) on the de-identification performances of the hybrid model. Both PDF extraction methods perform similarly in terms of token-level precision (P), recall (R), and F1-score.

Note that the edspdf method discards much more text than pdfbox. These metrics are consequently computed on texts that depend on the extraction technique (i.e., edspdf or pdfbox). This effect explains why the "fully redacted" metric is significantly higher when evaluating on documents extracted with "edspdf" instead of "pdfbox" although no difference is observed when considering the token-level performances.

| PDF extraction | P | R | F1 | Redacted | Fully redacted |
|---|---|---|---|---|---|
| **edspdf** | 99.1 ± 0.1 | 98.8 ± 0.1 | 98.9 ± 0.1 | 99.2 ± 0.1 | 93.1 ± 1.0 |
| **pdfbox** | 99.1 ± 0.0 | 98.9 ± 0.2 | 99.0 ± 0.1 | 99.4 ± 0.1 | 75.7 ± 3.0 |

Table 7: Performances of the de-identification on documents extracted with edspdf and pdfbox PDF extraction methods.

## Technical resources

The models in this study were trained on Nvidia V100 graphics cards, with each of the 110 experiments lasting approximately 40 minutes. CO2 equivalent emissions were estimated with CarbonTracker [22] at 113.40 gCO2eq for the NER finetuning of a single model for 4000 steps. If we include the emissions from fine-tuning the embedding model for the masked language modeling task, the total cost amounts to about 10 kgCO2eq [20].

# IV. Discussion

The work described in this article has allowed us to build a very efficient clinical text de-identification tool, which has been put into production within the AP-HP data warehouse with a daily analysis (about 50 000 documents per day). Out of 12 items to be extracted, 10 get a recall higher than 98%, which corresponds to the best results reported so far [23,1], although on datasets and languages that are difficult to compare.

Another important result of our work is a set of clinical texts in French, annotated with identifying information. Although it is not possible to share this dataset publicly (see the sharing conditions below), our different experiments allow us to share some lessons learnt and findings of interest to a team that would seek to create a powerful deidentification system.

# Lessons learnt about the annotation and the dataset

## Size of the training dataset

Although our own dataset contains more than 3600 annotated documents, our experiments with varying the size of the training set led us to the conclusion that excellent performance is achieved as early as 500 annotated documents, and that performance stops increasing significantly beyond 1000 documents (Figure 5).
Note that 10 documents only allow to reach as high as 0.91 of redacted metric.
These results suggest that with good quality annotation, with double annotation and double pass to ensure quality, a few hundred documents are sufficient to train a system of sufficient quality. Note that the first annotation phase with three annotators lasted 6 days at a rate of 7 hours per day, and the second phase lasted approximately 42 person-hours, for a total of 168 person-hours for 3600 documents. As the annotation of identifying information does not require special medical expertise, the human effort is not prohibitive.

## Document types

The system is not very sensitive to the omission of a document type in the training set, with most of the experiments on document type ablation not showing significant differences (Table 7). While it is always possible that documents with a very particular format could be mishandled by a generalist model, our results suggest that a random selection of documents is appropriate.

# Lessons learnt about the language model and the preprocessing

## Language model

The improvement brought by fine-tuning the language model (CamemBERT in our case) on the warehouse data is significant, which confirms many other works on the subject (Table 6). On the other hand, a complete retraining, which is extremely costly and time-consuming, does not bring any gain and is therefore not necessary.

## Text extraction step

The way to extract the text from the database (in our case, text extraction from PDF documents) matters a lot. In our case, a powerful preprocessing step removes up to 80% of the identifying information, and up to almost 100% for some types of entities, which benefits the robustness of the final system. However, because of the limited maturity of this extraction method and its gradual deployment in APHP's CDW, we decided to also include the documents extracted with the legacy PdfBox algorithm in the training set as well, in order to improve the end-to-end robustness of our document integration pipeline.

Including legacy documents in the training set also increased the number of identifying entities of all kinds, and most likely benefited the retrieval performance on these entities.Although this remark cannot be generalized to all systems, it remains true that it is impossible to separate the training of a named entity recognition system from the way the texts are stored and extracted.

### Tokenization

Since clinical texts often do not correspond to the standards on which the available tokenizers have been trained, it is necessary to pay attention to this step as well. For example, it is common for dates or measurements to be pasted in (e.g., "12nov", "5ml"). It is therefore very useful to systematically separate numbers and letters in the preprocessing.

## Machine learning vs rule-based approach

Expectedly, the rules have excellent precision but their recall is markedly insufficient. Hybridization, which consists in maximizing the recall by considering the entities predicted by at least one method, helps with a small number of annotated documents. However, Table 4 shows that overall, the hybridization improves the results only marginally, and that only two entity types (Visit id and NSS) give a better F1 score for rules alone compared to machine learning alone.

## Language specificities

Even if the system we describe here is only applicable to the French language, the work itself has only few language-related specificities.

First of all, the annotated corpus must of course be in the target language. This limitation is theoretical since, unfortunately, the confidentiality of the documents prevents sharing beyond a hospital group, which is an even stronger constraint.

Second, using a language-specific language model improves the results and allows domain-specific fine-tuning. Even if more and more language-specific BERT or equivalent models have been trained and shared in recent years, the coverage is far from complete and multilingual models are not always satisfactory compromises [24].

The exact same comment applies to the extraction and normalization of dates from the texts, for which systems exist in only some languages, and multilingual tools can be insufficient [25].

Note that in the case of a first-time de-identification process, using a fine-tuned or re-trained language model involves training on non-deidentified data. Therefore, this language model cannot be shared for multi-center projects, including within the institution [10]. Providing a model for downstream tasks thus requires re-training on the de-identified data, which doubles the cost in computing time and carbon (between several hours for fine-tuning and several days for retraining with 8 GPUs in our case [20]).

## Domain specificities

The medical field is far from being the only one for which the issue of de-identification of unstructured texts is crucial. Other domains such as legal texts or public administrations can benefit from the insights of this work, even if the documents are different in nature and some points are specific [26,27].

The code we provide with this article uses data in OMOP format, which focuses on medical data only [14]. This makes it easier to adapt to other data warehouses using the same data model. However, only the extraction related to structured data (patient name, etc.) and the *assign-and-replace* procedure are impacted by the format, which is not decisive.

More importantly, clinical reports can be quite distant from more controlled texts, in terms of syntax (many enumerations of noun groups without punctuation), spelling or structure (sections, line breaks due to extraction of text from PDF files, etc.).

Random date shifting is probably a useful feature in all domains, although the particular CNIL recommendation regarding differentiation between date of birth and date of clinical events is specific to medical reports [28]. Note also that most of our identifying entities are noun phrases, which is probably common in other fields as well.

## Maintenance and evolution

The performance of NER models degrades over time, due to several factors related to the change in the statistical properties of the variables of interest [29]. This may be the consequence of gradual changes in vocabulary, document distribution, and language habits. A re-evaluation of the system on recent documents should therefore be done regularly, for example every 3 years.

Note that even if our study did not focus on these considerations, attention should also be paid to more abrupt changes in information systems, input or pre-processing tools, which can have a more drastic negative effect. It is therefore necessary to check the de-identification process as part of the daily database quality checking process, by monitoring the quantity and distribution statistics of extracted and replaced entities.

## Data and model sharing

While we cannot publicly distribute the data and models described in this article, we are able to share the following resources that we hope will help new actors build their own de-identification systems:

- Source code used for building the dataset, training and applying the models, is freely available on AP-HP's Github account, distributed under a 3-Clause BSD license. It is documented, versioned and citable through Zenodo.
- The results of our experiments will help the community to optimize its annotation process and implementation choices based on our own experience.

Regarding the data and models themselves, access to the Clinical Data Warehouse's raw data can be granted following the process described on its website: eds.aphp.fr. Prior validation of the access by the local IRB is required. In the case of non-AP-HP researchers, the signature of a collaboration contract is also mandatory.

It is difficult, if not impossible, to share personal clinical texts due to strict regulations such as the General Data Protection Regulation (GDPR[3]). In addition, model transfer from one institution to another leads to performance losses [30], and many types of models, especially those based on neural networks, are likely to reveal sensitive information present in the training data [10,11]. Even if privacy-preserving methods have been suggested [31,32,33], performance is not yet close to the results obtained through traditionally supervised machine learning models.

These are both a limitation and the rationale for the process and results we share here, to facilitate the reproduction of a system of de-identification.

However, attacking machine learning models for re-identification is a complex task that requires a deep understanding of the model's underlying algorithm, training data, and implementation. While it is true that some studies have demonstrated the feasibility of attacking certain types of ML models, these attacks are often highly specialized and require significant resources to carry out. By implementing a provision behind an application programming interface (API) that requires authentication and limits the number of requests, it is possible to achieve a reasonable level of security against these attacks.

# Error analysis

---

[3] https://gdpr-info.eu/

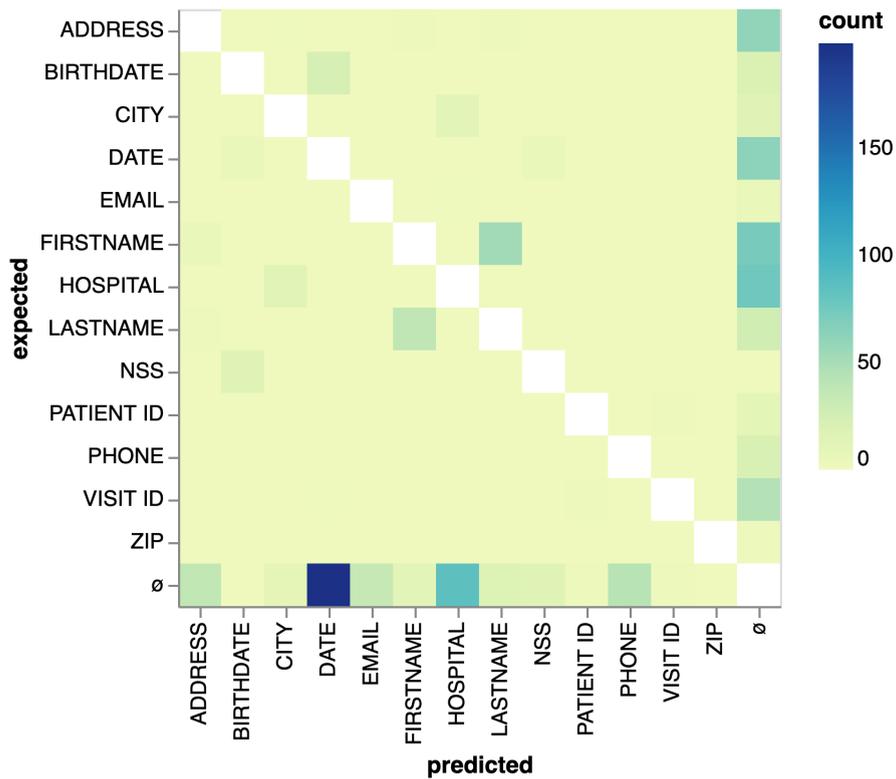

Figure 6: Confusion matrix highlighting the main sources of token-level error in the model's predictions

We performed an error analysis to better understand the types of errors the model is making. We provide a detailed breakdown of the token-level errors in the hybrid model's predictions through a confusion matrix (Figure 6). We also reviewed 143 tagging errors on the development set, and we propose a manual classification of these errors, distinguishing between false negatives (Figure 7) and false positives (Figure 8).

31% of the false positives concern medical information, which leads to a potential loss of information due to over-deidentification. For example, the model confuses the medical device "Joyce One", as well as the homeopathic medicine "Gelsur", the name of a temporary service "Covid" and "Arnold's neuralgia", with people.

The other wrongly predicted entities are less problematic as they cover team names, meaningless characters, non-medical common names or non identifying internal codes.

We have evaluated the potential cause of the entities missed by the model. We observe that in one third of the cases, the error is caused by the BERT tokenizer merging some years with the punctuation that follows them as in the example: "[ patient] [ reviewed] [ recently] [ (] [2007)]". These errors can be fixed by customizing the tokenizer to split around punctuation tokens. Other errors consist of rare formats such as dates that may be confused with floating point numbers "last meeting: 12.03", names separated by punctuation like "tom/smith", or missing spaces "Tom SmithSURGEON". Some errors were caused by names that are also common names, ambiguous acronyms such as the letter M in "M Smith".

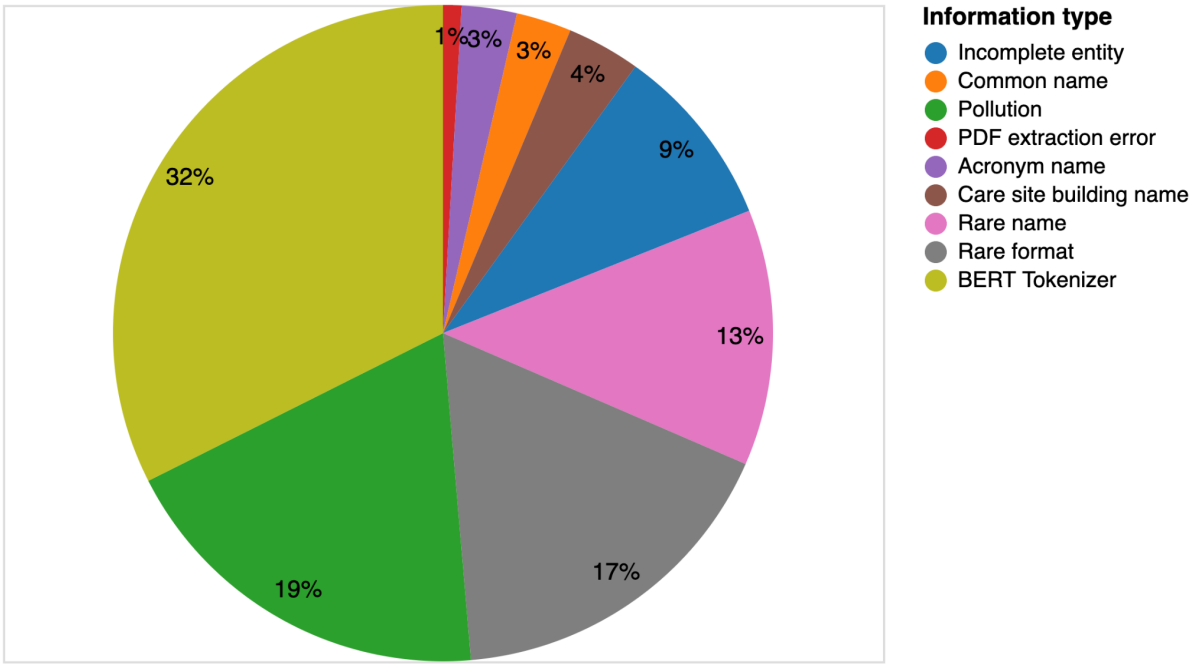

Figure 7: Main sources of errors in the entities that were missed entirely by the model

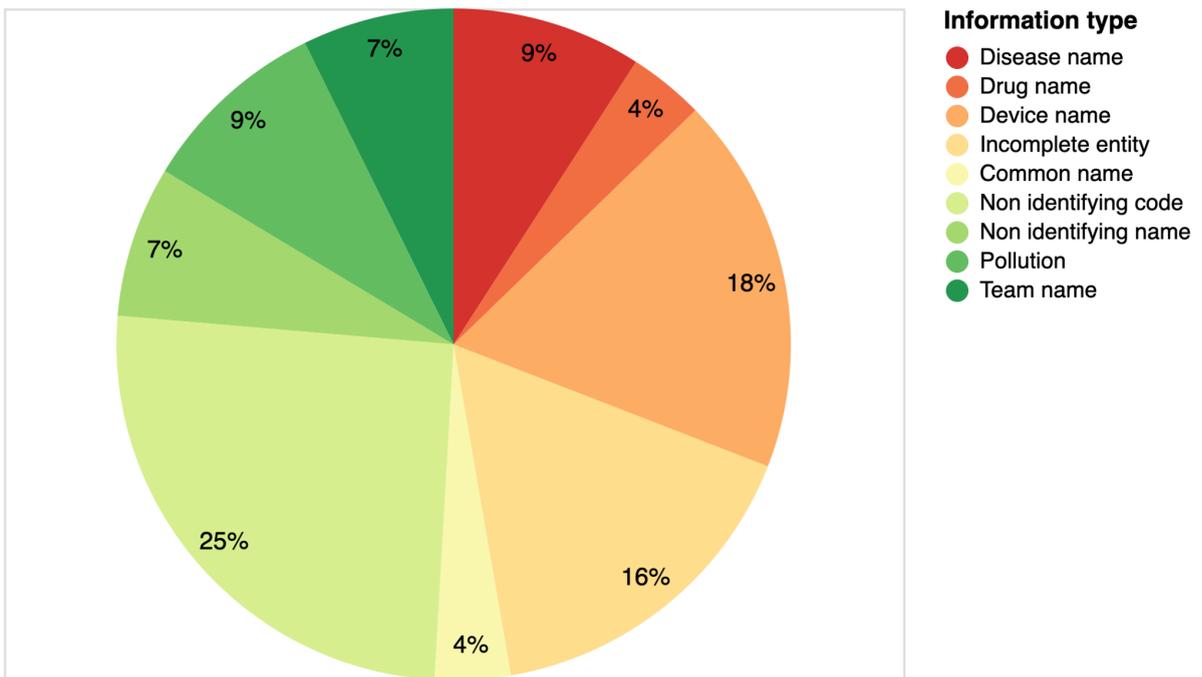

Figure 8: Coarse classification of fully false positive entities, i.e. entities that did not overlap any identifying entities of the test set.

# V.  Conclusion

This article has provided insights into the experience of the AP-HP team in implementing a de-identification system for medical documents. We have described the system and shared the resources, such as code and rule-based extraction patterns, that were used to achieve our results. We have discussed various implementation choices, including the size and types of documents in the dataset, as well as the benefits of fine-tuning a specific language model or adding static or dynamic rules. Furthermore, we have highlighted important considerations in pre-processing, computational cost, and carbon footprint, as well as the unique challenges of de-identification in the context of medical documents. By sharing our findings, we hope to provide guidance for others looking to implement efficient and effective de-identification systems, as well as contribute to the broader goal of protecting privacy in the context of medical data.

# Acknowledgment

We thank the clinical data warehouse (Entrepôt de Données de Santé, EDS) of the Greater Paris University Hospitals for its support and the realization of data management and data curation tasks.

# Authors contribution

All authors designed the study.
X.T.  drafted the manuscript. All authors interpreted data and made critical intellectual revisions of the manuscript.
X.T. did the literature review.
P.W. checked all the annotations.
P.W. , A.C. and B.D. developed the de-identification algorithms.
P.W. conducted the experiments and computed the statistical results.
X.T., A.M., M.H. and R.B. supervised the project.

# Conflict of Interest Disclosures

None reported.

# Data sharing

Access to the clinical data warehouse's raw data can be granted following the process described on its website: [eds.aphp.fr](https://eds.aphp.fr). Prior validation of the access by the local institutional review board is required. In the case of non-APHP researchers, the signature of a collaboration contract is moreover mandatory.

# Funding/Support

This study has been supported by grants from the AP-HP Foundation.

# Role of the Funder/Sponsor

The funder was involved neither during the design and conduct of the study nor during the preparation, submission or review of the manuscript.

# Supplementary Material

## A. STROBE & RECORD check list

|  | Item No. | STROBE items | Location in manuscript where items are reported | RECORD items | Location in manuscript where items are reported |
|---|---|---|---|---|---|
| **Title and abstract** | | | | | |

|  |  | (a) Indicate the study's design with a commonly used term in the title or the abstract (b) Provide in the abstract an informative and balanced summary of what was done and what was found | abstract | RECORD 1.1: The type of data used should be specified in the title or abstract. When possible, the name of the databases used should be included. | abstract |
|---|---|---|---|---|---|
|  | 1 |  |  | RECORD 1.2: If applicable, the geographic region and timeframe within which the study took place should be reported in the title or abstract. | abstract |
|  |  |  |  | RECORD 1.3: If linkage between databases was conducted for the study, this should be clearly stated in the title or abstract. | None |
| **Introduction** | | | | | |
| Background rationale | 2 | Explain the scientific background and rationale for the investigation being reported | introduction | | |
| Objectives | 3 | State specific objectives, including any prespecified hypotheses | | | |
| **Methods** | | | | | |
| Study Design | 4 | Present key elements of study design early in the paper | Methods | | |
| Setting | 5 | Describe the setting, locations, and relevant dates, including periods of recruitment, exposure, follow-up, and data collection | Methods | | |
| Participants | 6 | *(a) Cohort study - Give the eligibility criteria, and the sources and methods of selection of participants. Describe methods of follow-up*<br>*Case-control study - Give the eligibility criteria, and the sources and methods of case ascertainment and control selection. Give the rationale for the choice of cases and controls*<br>*Cross-sectional study - Give the eligibility criteria, and the sources and methods of selection of participants*<br><br>*(b) Cohort study - For matched studies, give matching criteria and number of exposed and unexposed*<br>*Case-control study - For matched studies, give matching criteria and the number of controls per case* | Methods | RECORD 6.1: The methods of study population selection (such as codes or algorithms used to identify subjects) should be listed in detail. If this is not possible, an explanation should be provided.<br><br>RECORD 6.2: Any validation studies of the codes or algorithms used to select the population should be referenced. If validation was conducted for this study and not published elsewhere, detailed methods and results should be provided.<br><br>RECORD 6.3: If the study involved linkage of databases, consider use of a flow diagram or other graphical display to demonstrate the data linkage process, including the number of individuals with linked data at each stage. | Methods/Data<br><br>None<br><br>None |
| Variables | 7 | Clearly define all outcomes, exposures, predictors, potential confounders, and effect modifiers. Give diagnostic criteria, if applicable. | NA | RECORD 7.1: A complete list of codes and algorithms used to classify exposures, outcomes, confounders, and effect modifiers should be provided. If these cannot be reported, an explanation should be provided. | NA |
| Data sources/ measurement | 8 | For each variable of interest, give sources of data and details of methods of assessment (measurement). Describe comparability of assessment | Methods | | |
| | | methods if there is more than one group | | | |
| Bias | 9 | Describe any efforts to address potential sources of bias | Methods (ablation studies) | | |
| Study size | 10 | Explain how the study size was arrived at | Methods (ablation studies) | | |
| Quantitative variables | 11 | Explain how quantitative variables were handled in the analyses. If applicable, describe which groupings were chosen, and why | NA | | |
| Statistical methods | 12 | (a) Describe all statistical methods, including those used to control for confounding<br>(b) Describe any methods used to examine subgroups and interactions<br>(c) Explain how missing data were addressed<br>(d) Cohort study - If applicable, explain how loss to follow-up was addressed<br>Case-control study - If applicable, explain how matching of cases and controls was addressed<br>Cross-sectional study - If applicable, describe analytical methods taking account of sampling strategy<br>(e) Describe any sensitivity analyses | Methods | | |
| Data access and cleaning methods | | | | RECORD 12.1: Authors should describe the extent to which the investigators had access to the database population used to create the study population.<br><br>RECORD 12.2: Authors should provide information on the data cleaning methods used in the study. | Methods<br><br>Methods |
| Linkage | | | | RECORD 12.3: State whether the study included person-level, institutional-level, or other data linkage across two or more databases. The methods of linkage and methods of linkage quality evaluation should be provided. | None |
| **Results** | | | | | |
| Participants | 13 | (a) Report the numbers of individuals at each stage of the study (e.g., numbers potentially eligible, examined for eligibility, confirmed eligible, included in the study, completing follow-up, and analysed)<br>(b) Give reasons for non-participation at each stage.<br>(c) Consider use of a flow diagram | Number of documents detailed in results | RECORD 13.1: Describe in detail the selection of the persons included in the study (i.e., study population selection) including filtering based on data quality, data availability and linkage. The selection of included persons can be described in the text and/or by means of the study flow diagram. | Number of documents detailed in results |
| Descriptive data | 14 | (a) Give characteristics of study participants (e.g., demographic, clinical, social) and information on exposures and potential confounders<br>(b) Indicate the number of participants with missing data for each variable of interest<br>(c) Cohort study - summarise follow-up time (e.g., average and total amount) | NA | | |
| Outcome data | 15 | Cohort study - Report numbers of outcome events or summary measures over time<br>Case-control study - Report numbers in each exposure category, or summary measures of exposure<br>Cross-sectional study - Report numbers of outcome events or summary measures | NA | | |
| Main results | 16 | (a) Give unadjusted estimates and, if applicable, confounder-adjusted estimates and their precision (e.g., 95% confidence interval). Make clear which confounders were adjusted for and why they were included<br>(b) Report category boundaries when continuous variables were categorized<br>(c) If relevant, consider translating estimates of relative risk into absolute risk for a meaningful time period | Results | | |
| Other analyses | 17 | Report other analyses done—e.g., analyses of subgroups and interactions, and sensitivity analyses | Ablation studies | | |
| **Discussion** | | | | | |
| Key results | 18 | Summarise key results with reference to study objectives | Discussion | | |
| Limitations | 19 | Discuss limitations of the study, taking into account sources of potential bias or imprecision. Discuss both direction and magnitude of any potential bias | Discussion | RECORD 19.1: Discuss the implications of using data that were not created or collected to answer the specific research question(s). Include discussion of misclassification bias, unmeasured confounding, missing data, and changing eligibility over time, as they pertain to the study being reported. | Discussion |
| Interpretation | 20 | Give a cautious overall interpretation of results considering objectives, limitations, multiplicity of analyses, results from similar studies, and other relevant evidence | Discussion | | |
| Generalisability | 21 | Discuss the generalisability (external validity) of the study results | Discussion | | |
| **Other Information** | | | | | |
| Funding | 22 | Give the source of funding and the role of the funders for the present study and, if applicable, for the original study on | Yes | | |

| | | which the present article is based | | | |
|---|---|---|---|---|---|
| Accessibility of protocol, raw data, and programming code | | | | RECORD 22.1: Authors should provide information on how to access any supplemental information such as the study protocol, raw data, or programming code. | Yes |

Table A: RECORD statement checklist

# Supplementary Results

## B. Data

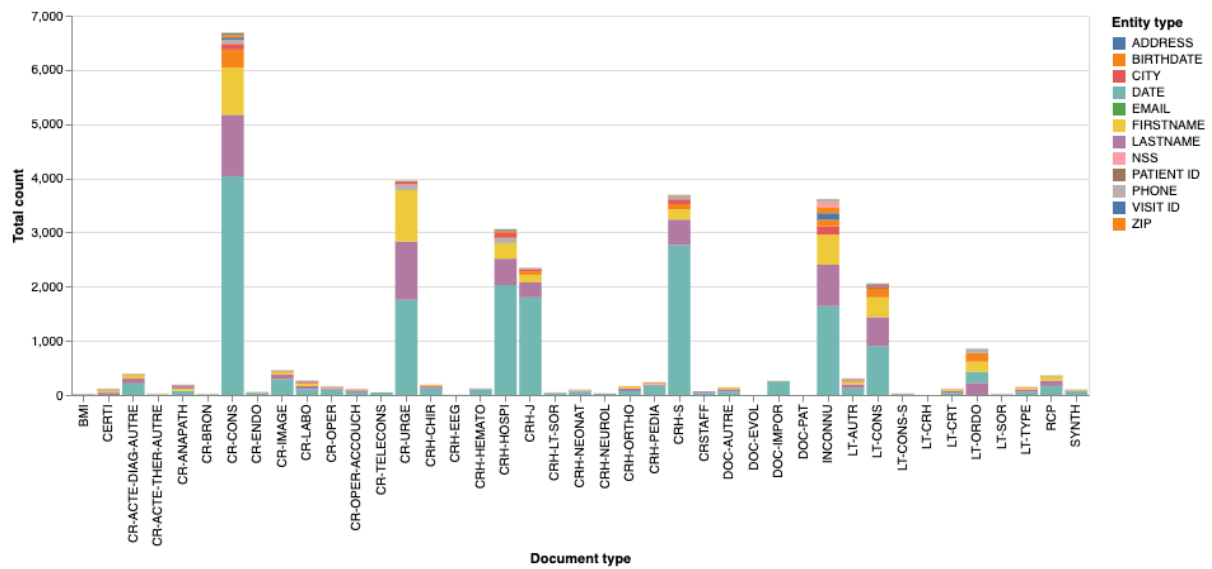

Figure B: Total number of annotated entities per document type, and per entity type. Only texts extracted with the edspdf method were considered.

We provide translations for the document types of Figure B and Figure 4:
- BMI: Initial evaluation note
- CERTI: Certificate
- CR-ACTE-DIAG-AUTRE: Diagnostic study note
- CR-ACTE-THER-AUTRE: Therapeutic evaluation note
- CR-ANAPATH: Pathology note
- CR-BRON: Bronchoscopy study (note)
- CR-CONS: Consult note
- CR-ENDO: Endoscopy study (note)
- CR-IMAGE: Diagnostic imaging study
- CR-LABO: Lab test results summary
- CR-OPER-ACCOUCH: Labor and delivery summary note

- CR-OPER: Surgery note
- CR-TELECONS: Outpatient consultation
- CR-URGE: Emergency department discharge summary
- CRH-CHIR: Surgery discharge summary
- CRH-EEG: EEG study
- CRH-HEMATO: Hematology discharge summary
- CRH-HOSPI: Hospital discharge summary
- CRH-J: Hospital discharge summary for day hospitalization
- CRH-LT-SOR: Discharge summary
- CRH-NEONAT: Neonatal perinatal medicine discharge summary
- CRH-NEUROL: Neurology discharge summary
- CRH-ORTHO: Orthopedic surgery discharge summary
- CRH-PEDIA: Pediatric discharge summary
- CRH-S: Hospital Discharge summary for week hospitalization
- CRSTAFF: Multidisciplinary comprehensive plan of care note
- DOC-AUTRE: Other document
- DOC-EVOL: Progress note
- DOC-IMPOR: Imported document
- DOC-PAT: Patient personnal document
- INCONNU: Unkown
- LT-AUTR: Other letter
- LT-CONS-S: Consult note
- LT-CONS: Consult note
- LT-CRH: Discharge summary
- LT-CRT: Miscellaneous letter
- LT-ORDO: Prescription list
- LT-SOR: Hospital letter
- LT-TYPE: Letter
- RCP: Multidisciplinary team meetings (MDTMs)
- SYNTH: Summary note

## C. Inter-annotator agreements

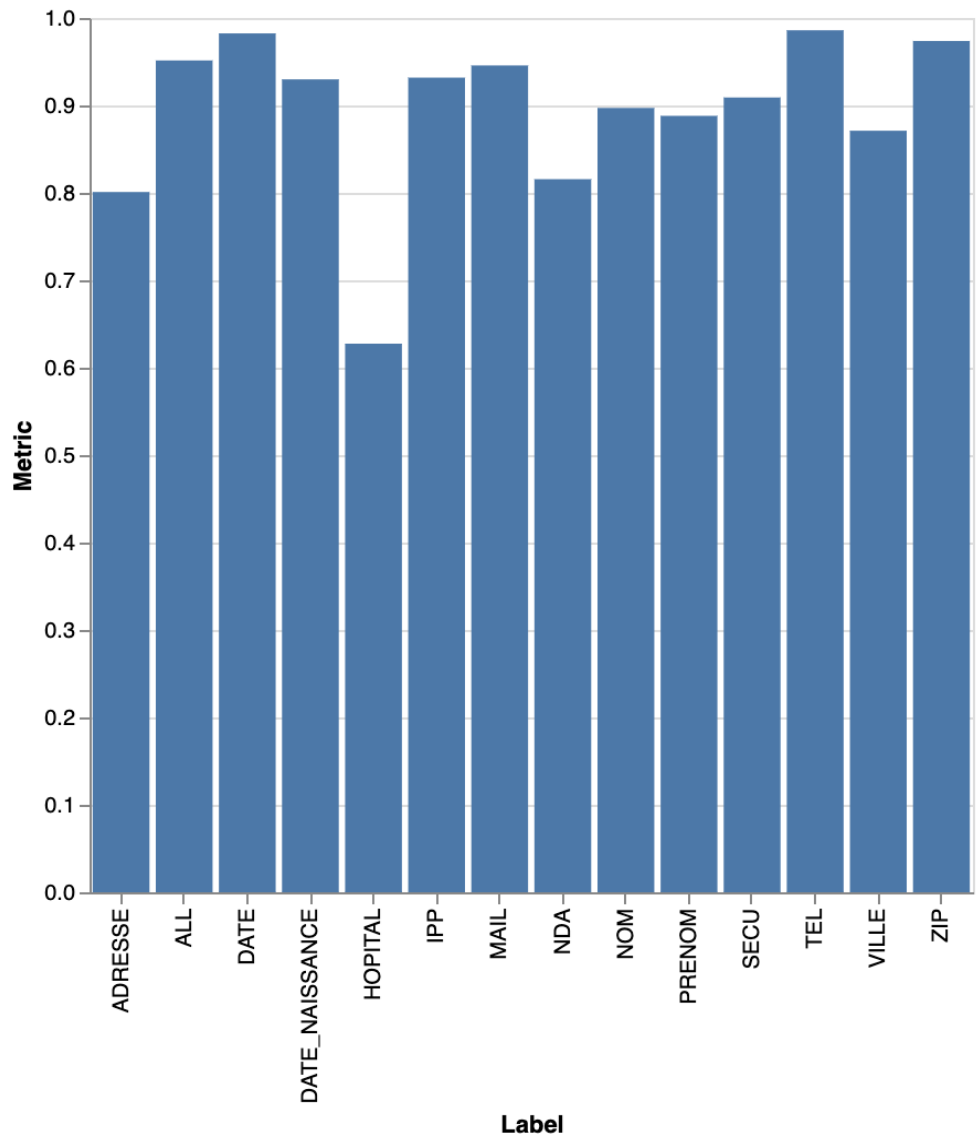

Figure C: Inter-annotator agreements (F1-score) by entity types.

## D. Rule-wise performance

|  | Static | | | Dynamic | | |
| --- | --- | --- | --- | --- | --- | --- |
|  | Precision | Recall | F1 | Precision | Recall | F1 |
| **ADDRESS** | 100.0 | 71.5 | 83.4 | 100.0 | 25.8 | 41.0 |
| **BIRTHDATE** | — | — | — | 100.0 | 0.0 | 0.0 |
| **CITY** | 100.0 | 26.5 | 42.0 | 100.0 | 39.8 | 57.0 |
| **DATE** | **93.0** | 97.0 | 94.9 | — | — | — |
| **EMAIL** | 100.0 | 94.2 | 97.0 | 100.0 | 0.0 | 0.0 |

| | | | | | | | |
|---|---|---|---|---|---|---|---|
| FIRSTNAME | | 98.3 | 22.5 | 36.7 | 98.3 | 22.2 | 36.2 |
| LASTNAME | | **82.5** | 45.0 | 58.2 | 98.8 | 28.0 | 43.6 |
| NSS | | 100.0 | 93.3 | 96.5 | 100.0 | 0.0 | 0.0 |
| PATIENT ID | | — | — | — | 100.0 | 96.2 | 98.0 |
| PHONE | | 100.0 | 95.5 | 97.7 | 100.0 | 0.1 | 0.2 |
| VISIT ID | | 100.0 | 14.7 | 25.6 | 100.0 | 35.3 | 52.2 |
| ZIP | | 100.0 | 66.7 | 80.0 | 100.0 | 55.6 | 71.4 |

Table D: final performances on the development set of the static and dynamic rules described in Table 2. These 12 rules were developed accessing both the training and the development sets. To build the final hybrid model, we discarded rules whose precision did not reach 98% (bold i.e., the joined FIRSTNAME/LASTNAME and the DATE static rules).

## E. Impact of the size of the training set

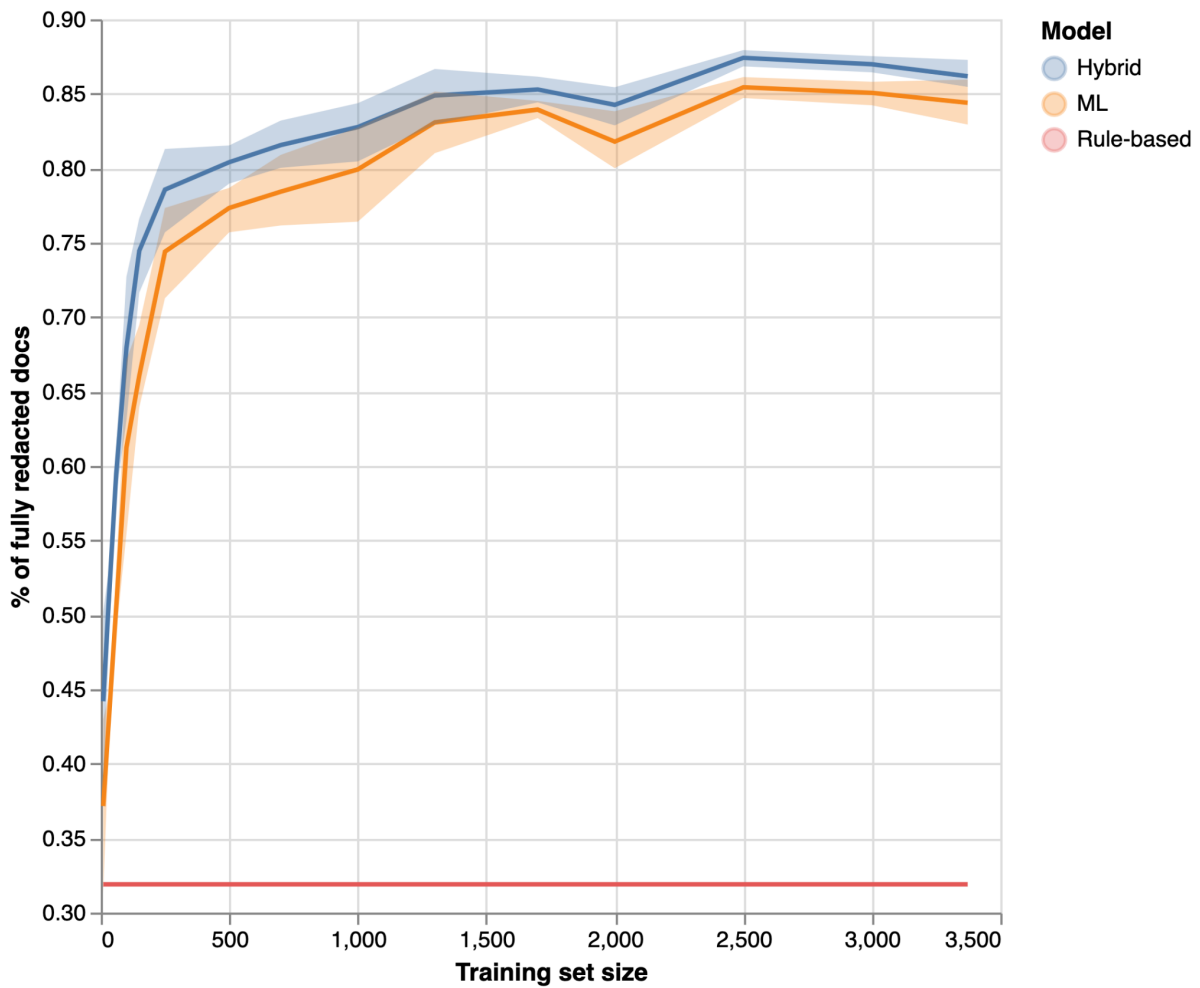

Figure E: Performance of the model on the test set in % of fully redacted documents for varying numbers of documents in the training set.

| Limit | Precision | Recall | F1 | Redacted | Fully redacted |
|---|---|---|---|---|---|
| **10** | 96.3 | 89.5 | 92.7 | 91.4 | 44.2 |
| **30** | 96.4 | 91.8 | 94.1 | 93.4 | 50.5 |
| **60** | 97.8 | 95.5 | 96.6 | 96.5 | 59.5 |
| **100** | 97.7 | 96.2 | 97.0 | 97.4 | 67.9 |
| **150** | 98.0 | 97.7 | 97.9 | 98.7 | 74.5 |
| **250** | 98.0 | 97.9 | 98.0 | 98.9 | 78.6 |
| **500** | 98.5 | 98.4 | 98.4 | 99.1 | 80.4 |
| **700** | 98.5 | 98.3 | 98.4 | 99.2 | 81.6 |
| **1000** | 98.8 | 98.6 | 98.7 | 99.3 | 82.8 |
| **1300** | 98.9 | 98.8 | 98.9 | 99.3 | 84.9 |
| **1700** | 98.9 | 98.9 | 98.9 | 99.4 | 85.3 |
| **2000** | 99.0 | 98.7 | 98.9 | 99.3 | 84.3 |
| **2500** | 98.9 | 99.0 | 98.9 | 99.5 | 87.4 |
| **3000** | 99.1 | 99.0 | 99.0 | 99.5 | 87.0 |
| **3373** | 99.0 | 98.9 | 99.0 | 99.4 | 86.2 |

Table E: performance of the hybrid model for various number of documents in the training dataset